\documentclass{article}

\usepackage{arxiv}

\usepackage[utf8]{inputenc} 
\usepackage[T1]{fontenc}    
\usepackage{hyperref}       
\usepackage{url}            
\usepackage{booktabs}       
\usepackage{amsfonts}       
\usepackage{nicefrac}       
\usepackage{microtype}      
\usepackage{lipsum}
\usepackage{graphicx}
\graphicspath{ {./images/} }

\usepackage{microtype}
\usepackage{graphicx}
\usepackage{subfigure}
\usepackage{booktabs} 

\usepackage{hyperref}

\usepackage{algorithmic}
\usepackage{algorithm}

\usepackage{amsmath}
\usepackage{amssymb}
\usepackage{mathtools}
\usepackage{amsthm}

\usepackage[capitalize,noabbrev]{cleveref}

\usepackage{import}
\usepackage{booktabs}
\usepackage{xcolor}
\usepackage{colortbl}
\usepackage{rotating}
\usepackage{multirow}
\usepackage[normalem]{ulem} 
\usepackage{rotating}
\usepackage{array}

\theoremstyle{plain}
\newtheorem{theorem}{Theorem}[section]

\theoremstyle{definition}

\newtheorem{assumption}[theorem]{Assumption}
\theoremstyle{remark}

\usepackage[textsize=tiny]{todonotes}

\title{TimEHR: Image-based Time Series Generation for\\
        Electronic Health Records}

\author{
 Hojjat Karami \\
  EPFL \\
  \texttt{hojjat.karami@epfl.ch} \\
   \And
 Mary-Anne Hartley \\
  Yale, EPFL\\
  \And
 David Atienza \\
  EPFL\\
  \And
 Anisoara Ionescu \\
  EPFL\\
}

\begin{document}
\maketitle
\begin{abstract}
Time series in Electronic Health Records (EHRs) present unique challenges for generative models, such as irregular sampling, missing values, and high dimensionality. In this paper, we propose a novel generative adversarial network (GAN) model, \textbf{TimEHR}, to generate time series data from EHRs. In particular, TimEHR treats time series as images and is based on two conditional GANs. The first GAN generates missingness patterns, and the second GAN generates time series values based on the missingness pattern. Experimental results on three real-world EHR datasets show that TimEHR outperforms state-of-the-art methods in terms of fidelity, utility, and privacy metrics.
\end{abstract}


\section{Introduction}
\label{intro}



Electronic health records (EHRs) chart patients' interactions with the health system and contain critical information for improving services and supporting research. Data from these systems are routinely incorporated into machine learning and statistical models for clinical decision support on diagnostic and prognostic predictions, as well as for monitoring health and evaluating treatment response \cite{tayefiChallengesOpportunitiesStructured2021}. However, access to large-scale EHR datasets is challenging and governed by strict regulations on privacy and security (e.g. HIPAA and GDPR), meaning that many models are based on unicentric data with a high risk of poor generalizability \cite{keshtaSecurityPrivacyElectronic2021}.

Traditional approaches for anonymization can be complex and costly, often compromising the data's statistical integrity and failing to provide robust privacy guarantees \cite{kushidaStrategiesDeidentificationAnonymization2012,azarm-daigleReviewCrossOrganizational2015}. The use of synthetic data is thus emerging as a promising solution for optimizing the trade-off between privacy and statistical utility \cite{appenzellerPrivacyUtilityPrivate2022,giuffreHarnessingPowerSynthetic2023}.

Generative models, particularly Generative Adversarial Networks (GANs) \cite{goodfellowGenerativeAdversarialNetworks2014}, have shown great potential in producing distribution-preserving synthetic EHR data. Their effectiveness spans various data modalities, including images \cite{singhMedicalImageGeneration2020,skandaraniGANsMedicalImage2023}, clinical text \cite{leeNaturalLanguageGeneration2018,huangClinicalBERTModelingClinical2020b}, discrete tabular EHRs (e.g. diagnoses and billing ICD codes) \cite{baowalySynthesizingElectronicHealth2019,torfiCorGANCorrelationCapturingConvolutional2020}, and time series \cite{yoonEHRSafeGeneratingHighfidelity2023,tianFastReliableGeneration2023}.

Time series data is a sequence of observations of one or multiple variables over a span of time. It can be regularly sampled (e.g., hourly temperature measurements) or irregularly sampled, such as spontaneous observation of clinical variables in the hospital. Sampling patterns introduce structured missingness that can be predictive of the outcome and thus create an important risk of `shortcut learning' (predicting outcomes from the presence/absence of data rather than its value). We can regard this as informative or biased sampling because clinicians order tests based on a prior probability that the patient has a certain condition, as drawn from their experience. In many real-world applications, missingness is not random and might depend on some unobserved factors (MNAR). However, many models assume that the missingness is completely random (MCAR) or missing at random (MAR) \cite{tanInformativeMissingnessWhat2023,getzenMiningEquitableHealth2023}.
Understanding these real-world patterns of missingness, is essential to differentiate signal from the noise.

The commonly used deep learning architectures for time series are Recurrent Neural Networks (RNNs) \cite{choiDoctorAIPredicting2016, choiRETAINInterpretablePredictive2017}, attention mechanism \cite{hornSetFunctionsTime2020,zhangGraphGuidedNetworkIrregularly2022}, or one-dimensional Convolutional Neural Networks (1D-CNNs) \cite{ramponiTCGANConditionalGenerative2018,liLearningIrregularlySampledTime2020,kosmaTimeParameterizedConvolutionalNeural2023a} where 1D convolutional operations are applied over the time dimension to capture temporal patterns and relationships. However, image-based time series modeling based on 2D-CNNs has also been employed for time series forecasting \cite{soodVisualTimeSeries2021,semenoglouImagebasedTimeSeries2023}, classification \cite{wangImagingTimeSeriesImprove2015, hatamiClassificationTimeSeriesImages2017,yazdanbakhshMultivariateTimeSeries2019}, and anomaly detection \cite{wenTimeSeriesAnomaly2019}. Especially for irregularly sampled time series, ViTST aims to transform irregularly sampled time series into line graphs and employ the Vision Transformer (ViT) for the subsequent classification task \cite{liTimeSeriesImages2023a}.

We propose a novel GAN-based model, \textit{TimEHR}, to generate synthetic EHR's time series data with irregular sampling and missing values. In particular, we treat each patient's time series as a 2-channel image (mask and values) and we use 2D-CNN architecture. TimEHR consists of two modules that are trained independently. The first module is a conditional Wasserstein GAN with gradient penalty (CWGAN-GP) that generates missing patterns (mask channel) from the noise and static data (e.g. demographics and outcome) as the conditional vector \cite{gulrajaniImprovedTrainingWasserstein2017}. The second module is a Pix2Pix GAN \cite{isolaImagetoImageTranslationConditional2017} that generates time series values from the missing pattern and the static data as the conditional vector. Once the two modules are trained, we can use pre-trained generators to generate time series data. The main contributions of this paper are as follows:

\begin{itemize}
    \item TimEHR \footnote{The code will be available on github after acceptance. for more information, please email the corresponding author.} can generate irregularly sampled time series with non-random missing values (MNAR). It is the first work to use image-based time series generation for EHRs.
    \item Our experiments on three large Electronic Health Record (EHR) datasets demonstrate that TimEHR outperforms state-of-the-art methods in terms of fidelity, utility, and privacy metrics.
    \item Our evaluation on simulated sinusoidal time series reveals that TimEHR is scalable to multivariate time series of up to length 128 and a number of variables up to 128, accommodating various missingness rates.
\end{itemize}

\section{Related Works}


Most works on time series generation focus on regularly sampled time series with no missingness. TimeGAN \cite{yoonTimeseriesGenerativeAdversarial2019a} is one of the early works that is based on an AutoEncoder and a GAN module (AE-GAN) with RNN architectures to obtain a dense latent space within which to generate samples. DoppelGANger \cite{linUsingGANsSharing2020} is another RNN-based model that also handles metadata in a conditional GAN framework. Here, we do not review the details of these works as they are not inherently compatible with irregularly sampled time series.

Generating irregularly sampled time series with missing values is an under-explored area.
T-CGAN \cite{ramponiTCGANConditionalGenerative2018} uses a conditional GAN framework with 1D-CNN architecture to generate irregularly sampled time series values conditioned on the timestamps vector. However, the timestamp vector is randomly generated and cannot handle missing values.
RTSGAN \cite{peiGeneratingRealWorldTime2021a} is a GRU-based AE-GAN model in which an observation embedding and a novel decide-and-generate decoder are proposed to handle irregular sampling and missingness.
GT-GAN \cite{jeonGTGANGeneralPurpose2022a} is also an AE-GAN method that utilizes various components, ranging from GRU-based neural ordinary/controlled differential equations to continuous time-ﬂow processes. However, it can only generate irregularly sampled time series without missing values.
EHR-Safe \cite{yoonEHRSafeGeneratingHighfidelity2023} is based on a two-stage model consisting of sequential encoder-decoder networks and GANs with stochastic normalization and explicit mask modeling ideas to improve utility and privacy.
Another line of research explores the use of diffusion models for time series generation.
TimeDIFF \cite{tianFastReliableGeneration2023} uses Denoising Diffusion Probabilistic Models (DDPM) and outperforms various real-world EHR datasets in terms of data utility and privacy.
TS-Diffusion \cite{liTSDiffusionGeneratingHighly2023} integrates the inhomogeneous Poisson process and observation probability into a diffusion model to overcome sample irregularity and missingness.

Among the mentioned works, RTSGAN, EHR-Safe, and TimeDIFF are the only models that can handle both irregular sampling and missingness and have been evaluated on EHR time series.











\section{TimEHR}

We begin by introducing the notation used in this paper. We then describe our proposed model in detail. 

\subsection{Problem Formulation}

An EHR dataset can be denoted as $\mathcal{D}=\{(s_i,T_i)\}_{i=1}^{N}$ where $N$ is the number of patients, $s_i$ is the static data (e.g. demographics, outcomes), and $T_i$ is the time series data such as vital signs and laboratory variables. The vector-based representation of each time series $T_i$ is denoted by $T_i=\{(t_j,\mathbf{x}_j,\mathbf{m}_j)\}_{j=1}^{L}$. Here, $L$ is the number of observations, $t_j \in \mathbb{R}_{\geq 0}$ is the time stamp, $\mathbf{x}_j \in \mathbb{R}^d$ is the vector of time series values ($d$ variables) and $\mathbf{m}_j \in \{0,1\}^d$ is the mask vector (1 if measured, 0 otherwise). The missing values in $\mathbf{x}_j$ are replaced by zero. Our objective is to generate a synthetic EHR dataset $\mathcal{D}_{syn}$ that is similar to the real EHR dataset $\mathcal{D}$. We make the following assumption:

\begin{assumption}
    We assume that the time series data $T_i$ causally depends on the static data $s_i$, i.e.,
    $P(s_i,T_i)=P(s_i)P(T_i|s_i).$
    \label{ass:independence}
\end{assumption}


Based on this assumption, instead of jointly generating static and time series data, we first use a tabular GAN to generate static data and then use the static data as the conditional vector to generate time series data using TimEHR. This is similar to RCGAN\cite{estebanRealvaluedMedicalTime2017a} and DoppelGANger\cite{linUsingGANsSharing2020} where the authors proposed to first generate metadata and then generate time series conditioned on the generated metadata.


\subsection{Time Series to Image}

\begin{figure}[!h]
    \centering
    \includegraphics[width=0.5\linewidth]{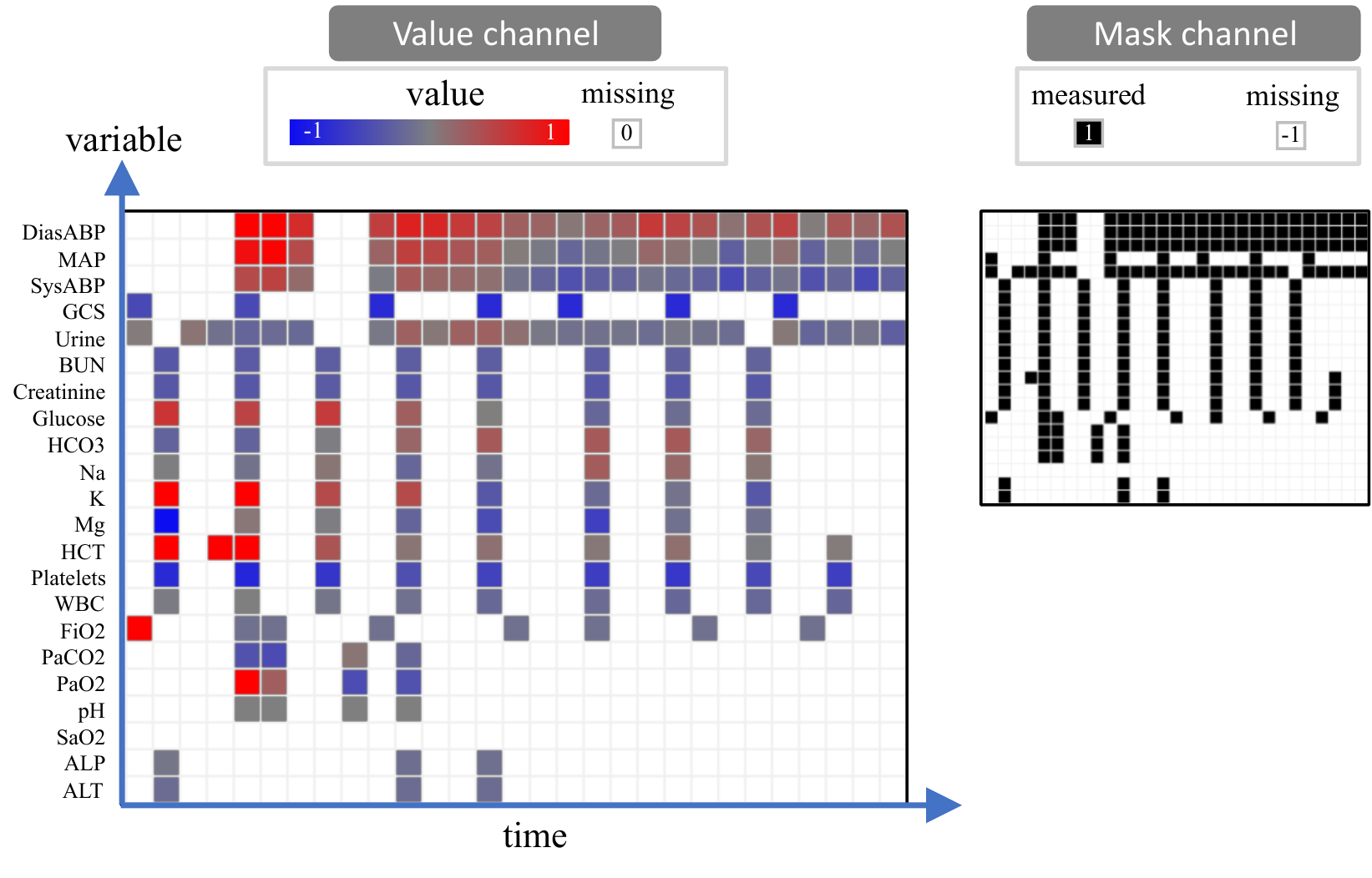}
    \caption{An image-based representation of a patient's time series data. Colors are for visualization purposes only.}
    \label{fig:sample}
\end{figure}

As we will treat time series data as images, we reshape each patient's time series data ($T_i$) into a two-channel image ($I_i$). For the first channel ($I_{i,value}$), we convert $\{(t_j,\mathbf{x}_j)\}_{j=1}^{L}$ into a matrix of shape $[d,H]$ where $d$ is the number of variables and $H$ is the number of time bins based on a desired time resolution ($r$). Here, $I_{i,value}[m,n]$ is the last available observation of $m$-th variable in the $\left((n-1)r,nr\right]$ interval (or zero if there is no observation). The value channel is min-max normalized to the [-1,1] range. The second channel ($I_{i,mask}$) is the corresponding mask matrix (-1 for missing values otherwise 1). It should be noted that our image is invariant to the order of variables and we can shuffle the variables to obtain a different image.
\cref{fig:sample} illustrates the image-view of an example patient's time series data. This is a compact form of representing time series data that highlights the non-random missingness pattern.

\subsection{Overall Architecture}

\begin{figure}[!h]
    \centering
    \includegraphics[width=0.4\linewidth]{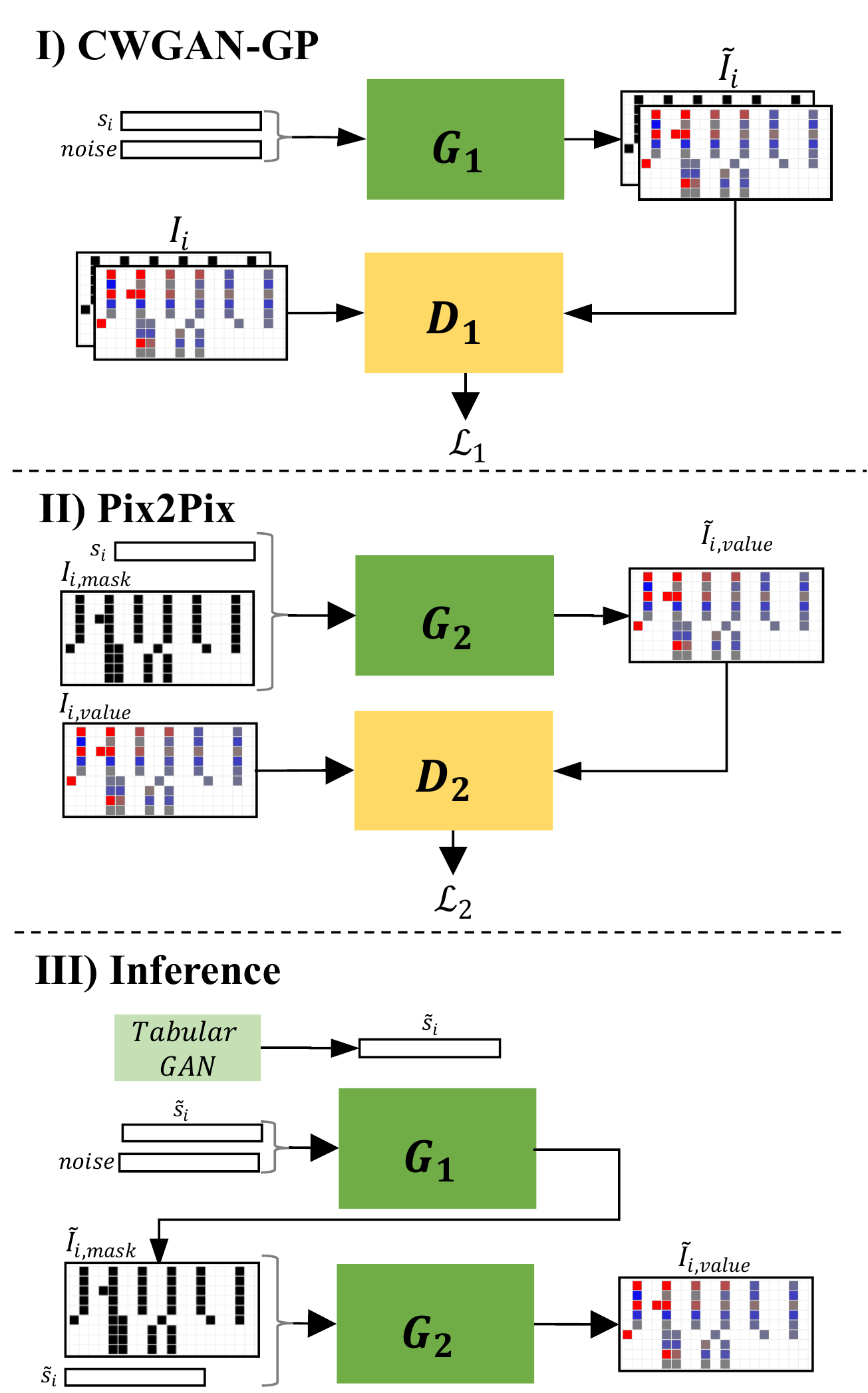}
    \caption{Model architecture. Module 1: CWGAN-GP for generating mask, Module 2: Pix2Pix for generating values, and Inference: generating synthetic time series.}
    \label{fig:arch}
\end{figure}

The overall architecture of TimEHR is shown in \cref{fig:arch}. The first module is a conditional Wasserstein GAN with Gradient Penalty (CWGAN-GP) that generates synthetic samples (mask and values) from noise and a conditional vector (static data and label). The second module is a Pix2Pix GAN \cite{isolaImagetoImageTranslationConditional2017} that generates synthetic samples (only values) from the conditional vector and the mask channel from the training data. In the inference step (\cref{fig:arch}-III), we use a Tabular GAN to generate static data, and the two trained generators to generate synthetic time series data.

\subsection{Module 1: CWGAN-GP}

We use a conditional Wasserstein GAN architecture with gradient penalty (CWGAN-GP) for generating images from noise and static data \cite{gulrajaniImprovedTrainingWasserstein2017}. The generator $G_1$ is a convolutional neural network (CNN) that takes as input a Gaussian noise vector $z$ ($z \sim \mathcal{N}(0, 1)$) and static data $s$, and outputs an image $\tilde{I}$. The discriminator $D_1$ is also a CNN that assigns scores for the generated and real images. The objective function of the CWGAN-GP is given by:
\begin{equation*}
    \begin{aligned} 
        \min_G \max_D V(G_1,D_1) =  \mathbb{E}_{I\sim P_{I}}[D_1(I,s)] \\
        - \mathbb{E}_{z\sim P_z}[D_1(G_1(z,s),s)]                      \\
        + \lambda_{GP} \mathbb{E}_{\hat{I}}\left[(\|\nabla_{\hat{I}}D_1(\hat{I},s)\|_2 - 1)^2\right],
    \end{aligned}
\end{equation*}

where $\hat{I} = \epsilon I + (1-\epsilon)G_1(z,s)$ and $\epsilon \sim U(0,1)$. The first two terms represent the Wasserstein distance or Earth Mover's Distance (EMD) between the distribution of real and generated data. The last term is the gradient penalty that enforces the Lipschitz constraint on the discriminator. $\lambda_{GP}$ is the gradient penalty coefficient.


\textbf{Architecture} The generator $G_1$ consists of a series of convolutional layers with batch normalization and ReLU activation. We apply an embedding layer to the static data $s$ and concatenate it with the noise vector $z$. The discriminator $D_1$ consists of a series of convolutional layers with batch normalization and LeakyReLU activation. We apply another embedding layer for the static data $s$ and add it as an additional channel to the input image.

\subsection{Module 2: Pix2Pix}
We use a conditional Pix2Pix architecture to generate the value channel from the mask channel \cite{isolaImagetoImageTranslationConditional2017}. The generator $G_2$ takes as input a mask channel ($I_{mask}$) and static data $s$ and outputs a value channel ($\tilde{I}_{value}=G_2(I_{mask},s)$). The discriminator $D_2$ takes as input a value channel ($I_{value}$ or $\tilde{I}_{value}$) and static data $s$ and outputs a scalar value $D_2(I_{value},s)$ that indicates the probability that the image is real. The objective function of the Pix2Pix is given by:
\begin{equation*}
    \begin{aligned}
        \mathcal{L}_{2}(G_2,D_2) = \mathbb{E}_{I_{value}}[\log D_2(I_{value},s)] \\
        + \mathbb{E}_{\tilde{I}_{value}}[\log (1-D_2(\tilde{I}_{value},s))].
    \end{aligned}
\end{equation*}

The first and second terms are the log probability the discriminator correctly classifies the real data as real and the generated data as fake, respectively. Our primary experiments showed that including noise as input to the generator does not help generate diverse value channels from a fixed mask channel which is consistent with the findings in \cite{mathieuDeepMultiscaleVideo2016,isolaImagetoImageTranslationConditional2017}.

Earlier methods have discovered that it is advantageous to combine the GAN objective with a reconstruction loss such as the L2 distance \cite{pathakContextEncodersFeature2016}. While the role of the discriminator remains the same, the generator is now assigned the dual task of deceiving the discriminator and closely matching the actual output in terms of L2 similarity. We explore this idea by adding L2 loss to the objective function:

\begin{equation*}
    \begin{aligned}
        \mathcal{L}_{rec}(G_2) = \mathbb{E}_{I_{value},\tilde{I}_{value}}[\|I_{value}-\tilde{I}_{value}\|_2].
    \end{aligned}
\end{equation*}

Our final objective is:
\begin{equation*}
    \begin{aligned}
        \min_{G_2} \max_{D_2} V(G_1,D_1) = \mathcal{L}_{2}(G_2,D_2) + \lambda \mathcal{L}_{rec}(G_2).
    \end{aligned}
\end{equation*}


\textbf{Architecture}
The generator has a U-Net architecture with skip connections \cite{ronnebergerUNetConvolutionalNetworks2015} with LeakyReLU activation in the down-sampling path and ReLU activation in the up-sampling path.
The discriminator is a PatchGAN classifier similar to \cite{isolaImagetoImageTranslationConditional2017} that consists of several CNN layers of 2D convolutions, batch normalization and LeakyReLU activation. This discriminator attempts to classify each $P \times P$ patch in an image as real or fake, and only penalizes structure at the patch scale.

\subsection{Inference (Time Series Generation)}

Once the models in the previous steps are trained, we can use the trained generators ($G_1$ and $G_2$) to generate time series data from noise and static data. Based on \cref{ass:independence}, we are flexible to use any conditional tabular GAN such as CTGAN in \cite{xuModelingTabularData2019} to generate static data. As the distribution of the label is important in downstream evaluation, we use the label as a conditional to generate other static variables. While the CWGAN-GP can independently generate realistic time series with missing data, we use only the mask channel as the input to the $G_2$. The inference procedure is shown in \cref{alg:inference}.

\begin{algorithm}[!htb]
    \caption{Generating Synthetic Time Series}
    \label{alg:inference}
    \begin{algorithmic}
        \STATE {\bfseries Input:} Size $N$, Generators $G_1$ and $G_2$
        \STATE {\bfseries Output:} synthetic dataset $\mathcal{D}_{syn}=\{(\tilde{s}_i,\tilde{T}_i)\}_{i=1}^{N}$

        \STATE Generate $N$ static vectors $\{\tilde{s}_i\}_{i=1}^N$ using CTGAN
        \STATE $\mathcal{D}_{syn}=\{\}$
        \FOR{$i=1$ {\bfseries to} $N$}
        \STATE generate noise $z_i$ from $\mathcal{N}(0, 1)$
        \STATE Generate mask pattern using $G_1$:
        \STATE $\quad$ $\tilde{I}_{i,mask} = G_1(z_i,\tilde{s_i}).mask$
        \STATE Generate values from mask pattern using $G_2$:
        \STATE $\quad$ $\tilde{I}_{i,value} = G_2(\tilde{I}_{i,mask},\tilde{s_i})$
        \STATE Reshape $\tilde{I}_{i,mask}$ and $\tilde{I}_{i,value}$ into $\tilde{T}_i$
        \STATE $\mathcal{D}_{syn}.append((\tilde{s}_i,\tilde{T}_i))$
        \ENDFOR
    \end{algorithmic}
\end{algorithm}

\section{Experiments}

\subsection*{Datasets}
We use three publicly available EHR datasets to evaluate TimEHR: (1) Medical Information Mart for Intensive Care III (MIMIC-III) \cite{johnsonMIMICIIIFreelyAccessible2016}, (2) the Physionet Challenge 2012 (P12) \cite{silvaPredictingInHospitalMortality}, and (3) the Physionet Challenge 2019 (P19) \cite{reynaEarlyPredictionSepsis2020}. The overall missing rates are 80.37\%, 74.58\%, and 79.40\% for P12, P19, and MIMIC-III, respectively and the main characteristics of the datasets are shown in \cref{tab:datasets}. We employ 5-fold cross-validation to evaluate the models. Since the number of variables and observations is less than 64, we pad time series images with zero to have an image of size $[2,64,64]$ (refer to \cref{app:datasets} for more details on data preprocessing).

\begin{table}[htbp]
    \small
    \centering
    \caption{Datasets description}

\begin{tabular}{ccccc}
\toprule
\toprule
Dataset & $N$ & $d$ & $L_{max}$ & Outcome(Prevalence) \\
\midrule
MIMIC-III & 51k   & 48    & 48    & Mortality(16.1\%) \\
P12   & 12k   & 35    & 48    & Mortality(14.2\%) \\
P19   & 38k   & 32    & 62    & Sepsis (7.2\%) \\
\bottomrule
\bottomrule
\end{tabular}%

    \label{tab:datasets}%

\end{table}%

\begin{table*}[htbp]
    \centering
    \caption{Fidelity Metrics: ($\uparrow$) higher is better, ($\downarrow$) lower is better.}

\begin{tabular}{ccccccc}
\toprule
Dataset & Model & precision ($\uparrow$) & recall ($\uparrow$) & density ($\uparrow$) & coverage ($\uparrow$) & TCD ($\downarrow$) \\
\midrule
\multirow{3}[2]{*}{P12} & RTSGAN & 0.441(0.036) & 0.428(0.018) & 0.237(0.025) & 0.381(0.025) & 0.055(0.008) \\
      & TimEHR & \textbf{0.775(0.02)} & \textbf{0.651(0.05)} & \textbf{0.657(0.054)} & \textbf{0.782(0.024)} & \textbf{0.019(0.001)} \\
      & BL    & \textit{0.855(0.015)} & \textit{0.854(0.006)} & \textit{0.94(0.051)} & \textit{0.967(0.006)} & \textit{0.01(0.001)} \\
\midrule
\multirow{3}[2]{*}{P19} & RTSGAN & 0.633(0.048) & 0.602(0.015) & 0.407(0.053) & 0.581(0.031) & 0.031(0.019) \\
      & TimEHR & \textbf{0.791(0.026)} & \textbf{0.657(0.03)} & \textbf{0.814(0.104)} & \textbf{0.77(0.051)} & \textbf{0.021(0.002)} \\
      & BL    & \textit{0.852(0.006)} & \textit{0.849(0.003)} & \textit{0.959(0.037)} & \textit{0.964(0.005)} & \textit{0.007(0.001)} \\
\midrule
\multirow{3}[2]{*}{MIMIC-III} & RTSGAN & 0.687(0.041) & 0.539(0.022) & 0.488(0.066) & 0.611(0.047) & 0.052(0.007) \\
      & TimEHR & \textbf{0.762(0.027)} & \textbf{0.622(0.039)} & \textbf{0.759(0.111)} & \textbf{0.798(0.055)} & \textbf{0.035(0.001)} \\
      & BL    & \textit{0.778(0.052)} & \textit{0.719(0.079)} & \textit{0.642(0.192)} & \textit{0.839(0.074)} & \textit{0.03(0.004)} \\
\bottomrule
\end{tabular}%

    \label{tab:fidelity}%

\end{table*}%

\subsection*{Evaluation Metrics}

We need a reasonable embedding of each time series data ($T_i$) to compute metrics. For each variable, we compute the common statistics (mean, standard deviation, min, max) and the missing rate, and concatenate them into a vector ($\mathbf{e}_i \in \mathbb{R}^{5d}$).
We briefly describe the evaluation metrics below (more details on how these metrics are calculated are provided in \cref{app:metrics}).

\textbf{Fidelity}
We report precision, recall, density, and coverage (PRDC) metrics between the estimated manifold of real and generated data distributions \cite{naeemReliableFidelityDiversity2020}. Additionally, we define a new metric called \textit{Temporal Correlation Difference (TCD)}, which is the \textit{L1 norm} of the difference between the correlation matrices of the real and generated data normalized by $d(d-1)/2$. To compute each correlation matrix, we first forward-fill time series value of each sample ($T_i$) and then compute the correlation matrix of the concatenated matrix of all samples along the time axis. We neglect the pairs if at least one of the variables is missing.

\textbf{Utility}
Following the Train on Synthetic, Test on Real (TSTR) protocol \cite{estebanRealvaluedMedicalTime2017a}, we train a LightGBM Classifier on the synthetic data and evaluate its performance on the test data. We report the area under the receiver operating characteristic curve (AUROC) as well as the area under the precision-recall curve (AUPRC) due to the high class imbalance of the datasets \cite{mcdermottCloserLookAUROC2024}.

\textbf{Privacy}
(1) \textit{Membership inference attack (MIA)}: This metric measures the probability of data belonging to the training set \cite{yoonEHRSafeGeneratingHighfidelity2023}. We fit a K-Nearest Neighbors (kNN) classifier on the synthetic data and calculate the nearest distances of train and test splits to the fitted model. Any differences between the distributions of the nearest distances of train and test splits indicate that the synthetic data is not private. We report the Jensen-Shannon Divergence ($JSD_{MIA}$) between the two distributions, as well as the AUROC of a binary classifier that predicts whether a sample is from the training set or not ($AUROC_{MIA}$).

(2) \textit{Nearest Neighbor Adversarial Accuracy Risk (NNAA)}: This score measures the degree to which a generative model overfits the real training data, a factor that could raise privacy-related concerns \cite{yaleGenerationEvaluationPrivacy2020}. It is the difference between two adversarial accuracies, $AA_{test} (AA(\mathcal{D}_{gen},\mathcal{D}_{test}))$ and $AA_{train} (AA(\mathcal{D}_{gen},\mathcal{D}_{train}))$ (see \cref{app:metrics}).





\subsection*{Baselines}
We use RTSGAN \cite{peiGeneratingRealWorldTime2021a} as the baseline because it is the only model with available code for generating irregularly sampled time series with missingness. Although models such as TimeGAN can be adapted to irregularly sampled time series with missingness (e.g. by defining mask matrix as additional features), we refrain from this approach as it leads to very poor performance in prior works \cite{peiGeneratingRealWorldTime2021a,yoonEHRSafeGeneratingHighfidelity2023,tianFastReliableGeneration2023}.

In addition to computing the metrics for TimEHR and the RTSGAN, we report the metrics for the Train-Test pair as an ideal baseline (BL). This is a valid idea, as we can consider the test data as highly authentic synthetic data. In addition, it can reveal an expected value for each metric. We use a 5-fold train-test split and report the mean and standard deviation of the metrics.



\begin{figure*}[h]
    \centering
    \includegraphics[width=0.95\textwidth]{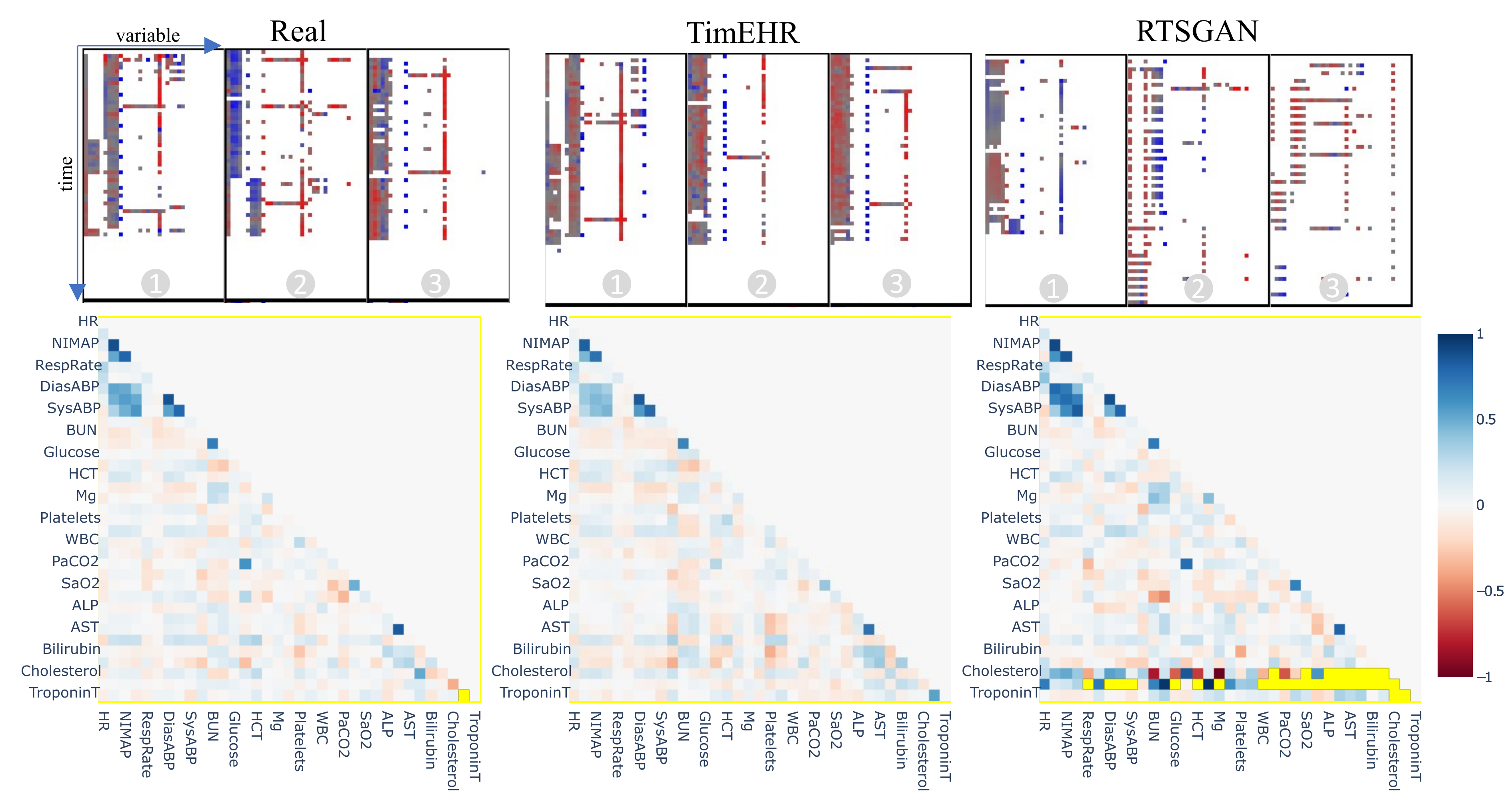}
    \caption{Top: Image-based visualization of three examples in P12. Bottom: Temporal Correlation comparison in P12. The NaN values are shown in yellow.}
    \label{fig:tempcorr}
\end{figure*}

\subsection{Overall Performance}

\textbf{Fidelity} \cref{tab:fidelity} shows the fidelity metrics for TimEHR and RTSGAN, as well as the test split (BL). We observe that TimEHR outperforms RTSGAN in terms of PRDC and TCD. Although the ideal value is typically 1 for PRDC and 0 for TCD, it is evident that this is not the case even for the test split (BL).

\cref{fig:tempcorr} illustrates image-based visualization of three examples (Top) and the correlation matrix (Bottom) from real data, TimEHR, and RTSGAN for the P12 dataset. Not only can TimEHR generate more realistic time series data than RTSGAN (especially the missingness pattern), but it can better preserve the temporal correlation structure of the variables. Notably, the RTSGAN correlation matrix has many NaN values (in yellow), which implies that many variable pairs (especially those with very high missingness rates) have not been generated together.
Although the input images to our model are invariant with respect to the order of variables, our model is still capable of capturing correlations between variables that are far from each other in the input image.

\begin{figure*}[h]
    \centering
    \includegraphics[width=0.8\textwidth]{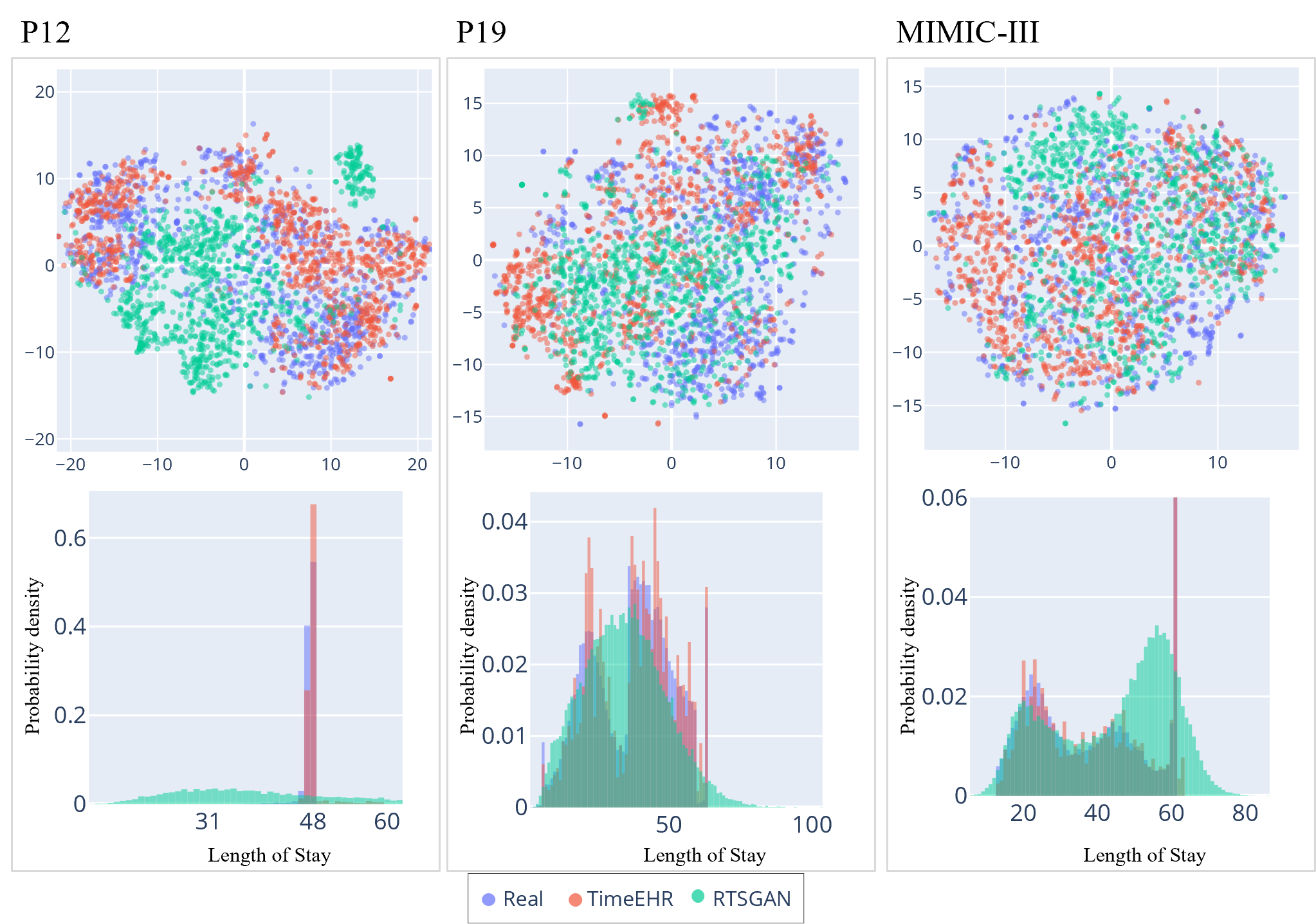}
    \caption{t-SNE visualization (top) and length of stay (bottom) for P12, P19 and MIMIC-III datasets.}
    \label{fig:tsne-los}
\end{figure*}

\cref{fig:tsne-los}-top illustrates the t-SNE visualization of the real data, and the generated data from TimEHR and RTSGAN for P12, P19, and MIMIC-III. We can see that the generated samples by TimEHR demonstrate a notably better overlap with the real data, which is also reflected in the PRDC metrics in \cref{tab:fidelity}. The histogram of the length of Stay (LOS) is also shown in \cref{fig:tsne-los}-bottom. Although TimEHR does not explicitly utilize any strategy for handling variable-length time series, it can generate realistic LOS distribution.








\textbf{Utility} \cref{tab:utility} shows the utility metrics for TimEHR and RTSGAN in the downstream task of binary classification. Here, we have also reported the Train on Real and Test on Real (TRTR) case, where we train a classifier on the training data and evaluate it on the test data. We see that TimEHR outperforms RTSGAN in terms of AUROC and AUPRC for all datasets. The performance gain is even higher in larger datasets (P19 and MIMIC-III), however, the AURPC is considerably lower than the TRTR case.


\begin{table}[!h]
    \small
    \centering
    \caption{Utility Metrics}
\begin{tabular}{cccc}
\toprule
Dataset & Model & AUROC & AUPRC \\
\midrule
\multirow{3}[2]{*}{P12} & RTSGAN & 0.745(0.015) & 0.331(0.03) \\
      & TimEHR & \textbf{0.785(0.015)} & \textbf{0.424(0.035)} \\
      & \textit{TRTR} & \textit{0.832(0.008)} & \textit{0.475(0.023)} \\
\midrule
\multirow{3}[2]{*}{P19} & RTSGAN & 0.677(0.037) & 0.141(0.022) \\
      & TimEHR & \textbf{0.853(0.014)} & \textbf{0.452(0.017)} \\
      & \textit{TRTR} & \textit{0.908(0.011)} & \textit{0.618(0.011)} \\
\midrule
\multirow{3}[2]{*}{MIMIC-III} & RTSGAN & 0.766(0.025) & 0.427(0.036) \\
      & TimEHR & \textbf{0.82(0.01)} & \textbf{0.536(0.024)} \\
      & \textit{TRTR} & \textit{0.863(0.012)} & \textit{0.621(0.028)} \\
\bottomrule
\end{tabular}%

    \label{tab:utility}%
\end{table}%

\textbf{Privacy} \cref{tab:privacy} shows the privacy metrics for TimEHR and RTSGAN. In general, none of the metrics shows any privacy issues. The AUROC and JSD metrics for MIA are close to the baseline for both models and show that the synthetic data is private. The adversarial accuracy (AA) metrics ($AA_{train}$, $AA_{test}$) are significantly higher than the baseline for both models. This is because these metrics are sensitive to mode collapse or mode generation, and any mismatch between the real and generated distributions can increase the adversarial accuracy (this can also be seen in PRDC metrics). Since the AA metrics do not show any privacy issues, we can conclude that the low NNAA metrics for both models only indicate the asymmetry between $AA_{train}$ and $AA_{test}$ rather than the privacy risk.


\begin{table*}[htbp]
    \small
    \centering
    \caption{Privacy Metrics. None of the metrics show privacy issues.}

\begin{tabular}{ccccccc}
\toprule
Dataset & Model & $AUROC_{MIA}$ & $JSD_{MIA}$ & $AA_{train}$ & $AA_{test}$ & $NNAA$ \\
\midrule
\multirow{3}[2]{*}{P12} & RTSGAN & 0.496(0.008) & 0.001(0.001) & 0.882(0.008) & 0.887(0.007) & -0.006(0.002) \\
      & TimEHR & 0.494(0.01) & 0.001(0.0) & 0.735(0.017) & 0.747(0.015) & -0.012(0.016) \\
      & \textit{BL} & 0.5   & 0     & 0.512(0.01) & 0.512(0.01) & 0 \\
\midrule
\multirow{3}[2]{*}{P19} & RTSGAN & 0.5(0.009) & 0.001(0.0) & 0.767(0.01) & 0.779(0.009) & -0.012(0.01) \\
      & TimEHR & 0.499(0.008) & 0.001(0.0) & 0.721(0.025) & 0.736(0.029) & -0.015(0.012) \\
      & \textit{BL} & 0.5   & 0     & 0.518(0.012) & 0.518(0.012) & 0 \\
\midrule
\multirow{3}[2]{*}{MIMIC-III} & RTSGAN & 0.491(0.025) & 0.005(0.002) & 0.777(0.021) & 0.856(0.057) & -0.079(0.04) \\
      & TimEHR & 0.488(0.024) & 0.004(0.002) & 0.733(0.041) & 0.837(0.061) & -0.104(0.062) \\
      & \textit{BL} & 0.5   & 0     & 0.669(0.092) & 0.669(0.092) & 0 \\
\bottomrule
\end{tabular}%

    \label{tab:privacy}%

\end{table*}%


\subsection{Ablation Study}

\begin{figure}[h]
    \centering
    \includegraphics[width=0.4\textwidth]{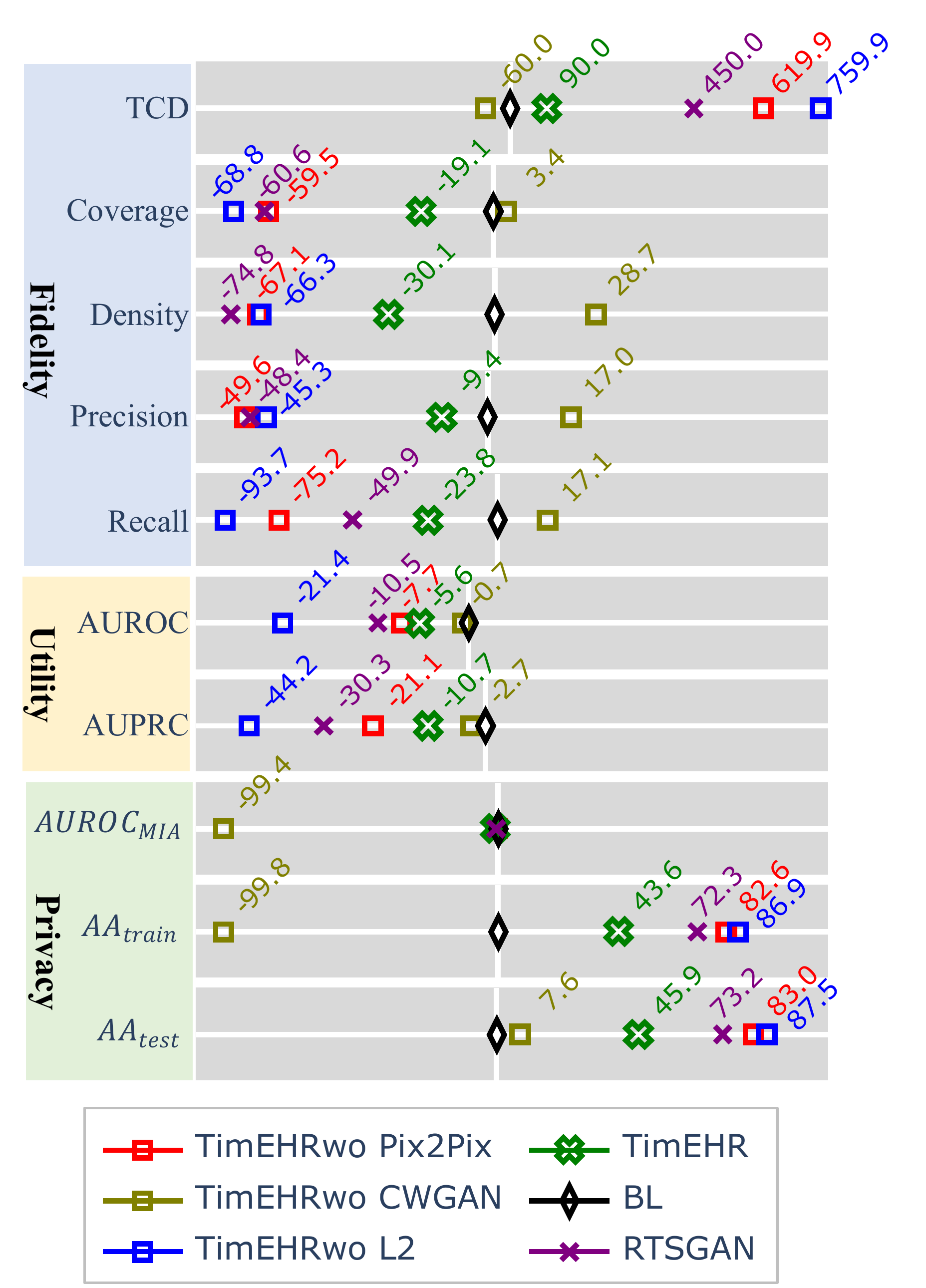}
    \caption{Ablation study. Values are percent deviation from the baseline.}
    \label{fig:ablation}
\end{figure}

We can further investigate whether the use of each of the proposed modules improves the performance of the model. We conducted an ablation study by removing each module and comparing the results with the full model. The percent deviation of each metric from its baseline is shown in \cref{fig:ablation}. We observe that \textit{TimEHR w/o Pix2Pix} (\textcolor{red}{\textbf{$\square$}}
) and \textit{TimEHR wo L2} (\textcolor{blue}{\textbf{$\square$}}
) have the worst fidelity and utility scores while preserving privacy. On the other hand, \textit{TimEHR w/o CWGAN} (\textcolor{olive}{\textbf{$\square$}}
) seems to have a perfect fidelity and utility score. However, low privacy metrics ($AUROC_{MIA}$ and $AA_{train}$) show that it is not a privacy-preserving model and it is memorizing the training data. This can be attributed to the deterministic nature of Pix2Pix module as it ignores the input noise. Additionally, we can argue that having fidelity scores that are better than the baseline is associated with privacy issues. Finally, \textit{TimEHR} (\textcolor{green}{\textbf{$\times$}}
) has the best performance in terms of fidelity and utility while preserving privacy. Based on these results, we can conclude that the CWGAN module improves the privacy of the model by generating the missingness pattern, and the Pix2Pix module with L2 reconstruction loss improves the fidelity and utility of the model by preserving the temporal correlation structure.

\subsection{Simulated Data}
\begin{figure}[h]
    \centering
    \includegraphics[width=0.4\textwidth]{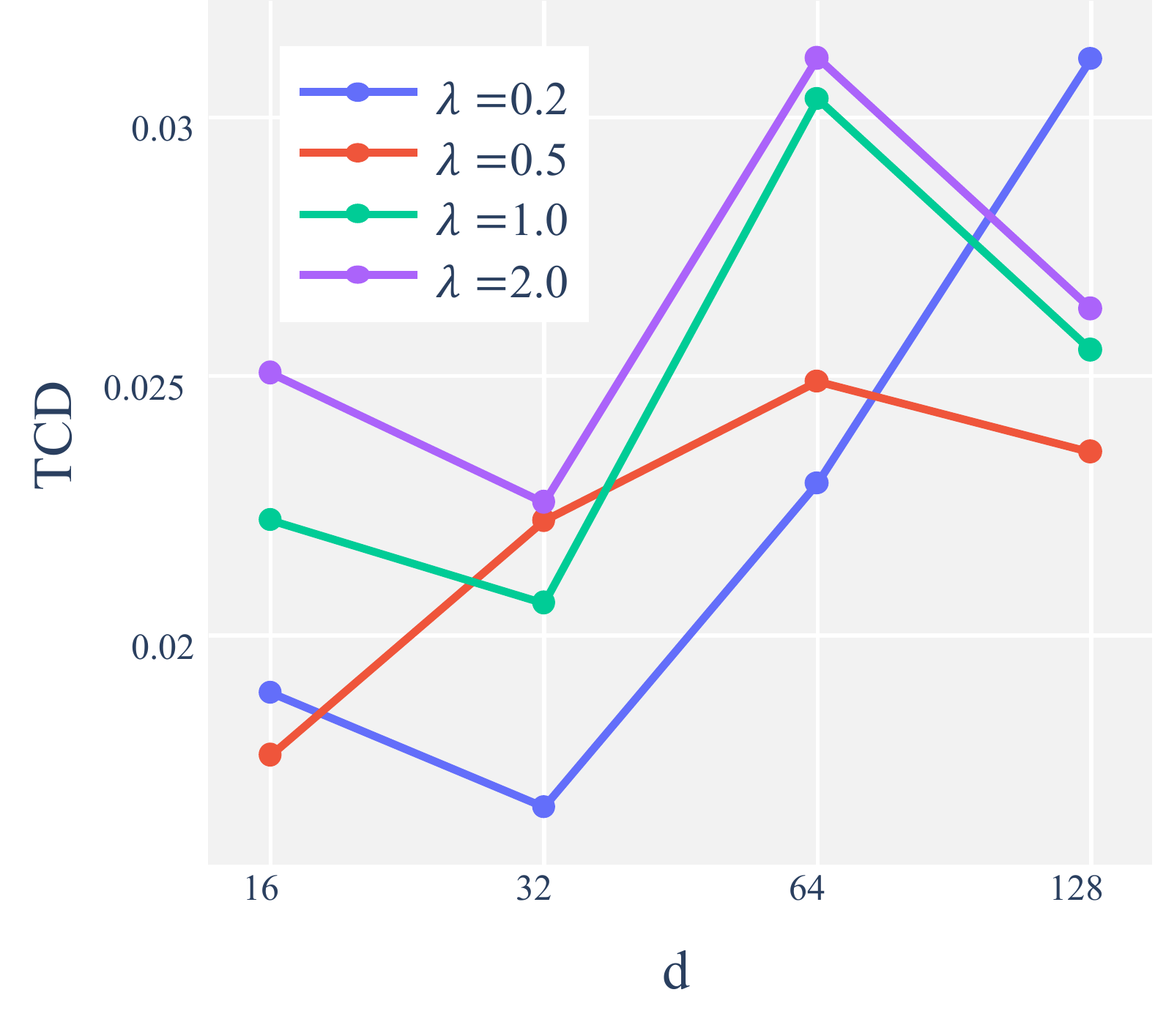}
    \caption{TCD for simulated data.}
    \label{fig:sim-tcd}
\end{figure}

We further investigate TimEHR on simulated sinusoidal time series data to understand its capabilities. We generate 10,000 time series with $L=64$ and $d \in \{16,32,64\}$, as well as a longer time series with $L=128$ and $d=128$. A $128 \times 128$ image seems to be a reasonable proxy for most of the EHR datasets as it can capture a time horizon of around 5 days (hourly data) and 128 variables. For each variable, $x_d(t) = A_{d} \sin(\frac{2\pi}{P_d}t + \phi_d) + \epsilon$ (full data generating process in \cref{app:datasets}). Additionally, the missing pattern follows a homogenous Poisson process with intensity function $\lambda^{*}(t)=\lambda$. We use $\lambda \in \{0.2,0.5,1,2\}$ which corresponds to an overall missing rate of  81.85\%, 60.66\%, 36.78\%, and 13.64\% respectively. \cref{fig:sim-tcd} shows the TCD metric for different number of variables ($d$) and values of $\lambda$. The maximum TCD value is still less than 0.035, which shows that TimEHR can preserve the temporal correlation structure of the variables even for a large number of variables and a high missing rate.

\section{Conclusion}
We have proposed TimEHR to generate irregularly sampled time series with missingness from EHR datasets. It shows promising results in terms of fidelity, utility, and privacy metrics on three large EHR datasets.

\textbf{Limitations and Future Works} Although our model exhibits acceptable privacy metrics, it is not a privacy-preserving model as we have not considered any privacy constraints in the training process. Thus, we leave it to future work to incorporate privacy-preserving frameworks such as differential privacy into the training process \cite{xieDifferentiallyPrivateGenerative2018a}. It is also interesting to explore other CNN-based architectures to see if they can improve the performance of our model in terms of model utility. In this work, we have experimented TimEHR on time series images of size $64 \times 64$ in EHR datasets and $128 \time 128$ in the simulated datasets. However, it would be interesting to see how the model performs on datasets with longer time series and more variables.

\textbf{Why we have used 2D-CNNs for time series?}
Our primary innovation lies in the utilization of the reshape operation, transforming the time series into a two-channel image. Retaining the missing pattern as an additional channel facilitates the CWGAN-GP module in generating realistic missingness patterns. Despite employing a kernel size of 4, the extensive layering in our CNN architecture ensures that the model's receptive field is sufficiently large to capture the temporal correlation structure of the variables.

\textbf{Is our model suitable for regularly sampled time series?} In this work, we have used the missingness pattern to generate realistic time series. As a result, our second module (Pix2Pix) is unuseful for regularly sampled time series, as it generates values based on the missingness pattern. However, the CWGAN module can still be used to generate the values. We leave it for future work to investigate the performance of our model on regularly sampled time series.

\bibliographystyle{unsrt}  

\bibliography{template}

\newpage

\section{Datasets}
\label{app:datasets}

In this work, we use three publicly available EHR datasets and a simulated dataset to evaluate TimEHR:

\subsection*{MIMIC-III}
Medical Information Mart for Intensive Care III (MIMIC-III) \cite{johnsonMIMICIIIFreelyAccessible2016} is a centralized database with an extensive set of electronic health record (EHR) data for individuals admitted to critical care units at Beth Israel Deaconess Medical Center from 2001 to 2012. The dataset encompasses details such as patient demographics, laboratory results, vital signs, procedures, caregiver notes, and outcomes. We used 48 time series and four static variables (age, height, weight, mortality) using the reproducibility pipeline from \cite{johnsonReproducibilityCriticalCare2017a}. This led to 50,961 patients. The overall missingness rate is 79.40\%.

\subsection*{Physionet Challenge 2012 (P12)}
This dataset \cite{silvaPredictingInHospitalMortality} contains 35 time series and six static variables (age, gender, height, weight, ICU type, and mortality) for 12,000 ICU patients. The overall missingness rate is 80.37\%.

\subsection*{Physionet Challenge 2019 (P19)}
This dataset \cite{reynaEarlyPredictionSepsis2020} consists of 32 time series and four static variables (age, gender, admission time, sepsis shock) for 40,333 patients. The overall missingness rate is 74.58\%.

\subsection*{Simulated Data}

The full algorithm for generating simulated data is shown in \cref{alg:sim}.




\begin{algorithm}[htb]
    \caption{Generating Simulated Time Series}
    \label{alg:sim}
    \begin{algorithmic}
        \STATE {\bfseries Input:} Size $N$, number of variables $d$, Poisson rate $\lambda$, Time series length $L$, Period range $[T_{min},T_{max}]$

        Data=np.zeros((N,L,d))

        \FOR{$d=1$ {\bfseries to} $D$}
        \STATE $\phi_d=uniform(0,2\pi)$
        \STATE $T_d=rand(T_{min},T_{max})$
        \FOR{$i=1$ {\bfseries to} $N$}
        \STATE Generate Values:
        \STATE $\quad$ $t = np.arange(L)$
        \STATE $\quad$ $A = U(0.9,1)$
        \STATE $\quad$ $b = U(-0.5, 0.5)$
        \STATE $\quad$ $\epsilon = Normal(0, 0.1, L)$
        \STATE $\quad$ $\epsilon_T = rand(1,5)$
        \STATE $\quad$ $\epsilon_T = U(0, 2\pi/50)$
        \STATE $\quad$ $Data[i,:,d] = b + A  sin(2  \pi  t / (T_d+\epsilon_T) + \phi_d) + \epsilon$
        \STATE Set Missingness Pattern:
        \STATE $\quad$ $poisson\_counts = Poisson(\lambda, L)$
        \STATE $\quad$ $mask = np.where(poisson\_counts > 0, 1, 0)$
        \STATE $\quad$ $Data[i,mask==0,d] = NaN$

        \ENDFOR
        \ENDFOR

    \end{algorithmic}
\end{algorithm}

\subsection*{Data Preprocessing}
We use a 5-fold train-test split. We use the first 64 (48) hours of each patient's ICU stay for P19 (P12 and MIMIC-III) and the time series are aggregated into 1-hour intervals. If we have multiple observations, we keep the last one and in the case of no observation we use NaN. Then, they are normalized to have zero mean and unit variance (NaN values are ignored), and the outlier values (more than 3 standard deviations from the mean) are clipped. Finally, we perform a min-max normalization to scale the values between -1 and 1. For static variables, we use one-hot encoding for categorical variables and standardization for continuous variables.

The next step is to reshape the time series into images. Each patient time series is a matrix of shape $[L_i, d]$ where $L_i$ is the length of the time series and $d$ is the number of variables. We pad this matrix with zeros to have a shape of $[64,64]$. We form a similar mask matrix where 1 indicates a measured value and -1 indicates a missing value. We replace NaN values with 0 in the time series matrix. Finally, we concatenate the time series and mask matrices along the channel axis to form a 2-channel image of shape $[2,64,64]$.

\section{Training Details}

The training algorithms for CWGAN-GP and Pix2Pix are shown in \cref{alg:cwgan} and \cref{alg:pix2pix}, respectively.

\begin{algorithm}[htb]
    \caption{Training CWGAN-GP}
    \label{alg:cwgan}
    \begin{algorithmic}
        \STATE {\bfseries Input:} Training data $\mathcal{D}_{train}$, Batch size $B$
        \STATE {\bfseries Initialize:} $G_1$ and $D_1$.

        \FOR{$epoch=1$ {\bfseries to} $n_{epochs}$}
        \STATE sample minibatch of $B$ noise samples $\{z_1,\ldots,z_B\}$ from $P_z$
        \STATE sample minibatch of $B$ real data samples $\{I_1,\ldots,I_B\}$ from $P_{data}$
        \STATE Training $D_1$:
        \STATE $\quad$ $\tilde{I}_i = G_1(z_i,s_i)$
        \STATE $\quad$ $\hat{I_i} = \epsilon I_i + (1-\epsilon)\tilde{I}_i$
        \STATE $\quad$ $\mathcal{L}_{D_1} = -\frac{1}{B}\sum_{i=1}^{B}D_2(I_i) + \frac{1}{B}\sum_{i=1}^{B}D_2(\tilde{I}_i)$
        \STATE $\quad$ $\mathcal{L}_{D_1} = \mathcal{L}_{D_1} + \lambda_{GP} \mathbb{E}_{\hat{I}\sim P_{\hat{I}}}[(\|\nabla_{\hat{I}}D_1(\hat{I})\|_2 - 1)^2]$
        \STATE $\quad$ Update $D_1$ using $\nabla_{D_1}\mathcal{L}_{D_1}$
        \STATE Training $G_1$:
        \STATE $\quad$ $\tilde{I}_i = G_1(z_i,s_i)$
        \STATE $\quad$ $\mathcal{L}_{G_1} = -\frac{1}{B}\sum_{i=1}^{B}D_1(\tilde{I}_i)$
        \STATE $\quad$ Update $G_1$ using $\nabla_{G_1}\mathcal{L}_{G_1}$

        \ENDFOR
    \end{algorithmic}
\end{algorithm}

\begin{algorithm}[htb]
    \caption{Training Pix2Pix}
    \label{alg:pix2pix}
    \begin{algorithmic}
        \STATE {\bfseries Input:} Training data $\mathcal{D}_{train}$, Batch size $B$
        \STATE {\bfseries Initialize:} $G_2$ and $D_2$.

        \FOR{$epoch=1$ {\bfseries to} $n_{epochs}$}
        \STATE sample minibatch of $B$ real data missing pattern $\{I_{1,mask},\ldots,I_{B,mask}\}$ from  $\mathcal{D}_{train}$
        \STATE Training $D_2$:
        \STATE $\quad$ $\tilde{I}_{i,value} = G_2(I_{i,mask},s_i)$
        \STATE $\quad$ $\mathcal{L}_{D_2} = -\frac{1}{B}\sum_{i=1}^{B}[\log D_2(I_{i,value},s_i)] - \frac{1}{B}\sum_{i=1}^{B}[\log (1-D_2(\tilde{I}_{i,value},s_i))]$
        \STATE $\quad$ Update $D_2$ using $\nabla_{D_2}\mathcal{L}_{D_2}$
        \STATE Training $G_2$:
        \STATE $\quad$ $\tilde{I}_{i,value} = G_2(I_{i,mask},s_i)$
        \STATE $\quad$ $\mathcal{L}_{G_2} = -\frac{1}{B}\sum_{i=1}^{B}[\log D_2(\tilde{I}_{i,value},s_i)]$
        \STATE $\quad$ $\mathcal{L}_{rec} = \frac{1}{B}\sum_{i=1}^{B}[\|I_{i,value} - \tilde{I}_{i,value}\|_2]$
        \STATE $\quad$ $\mathcal{L}_{G_2} = \mathcal{L}_{G_2} + \lambda \mathcal{L}_{rec}$
        \STATE $\quad$ Update $G_2$ using $\nabla_{G_2}\mathcal{L}_{G_2}$

        \ENDFOR
    \end{algorithmic}
\end{algorithm}

\subsection*{Hyperparameters}

We used Optuna library \cite{akibaOptunaNextgenerationHyperparameter2019} for hyperparameter tuning. We explored learning rate in the range of $[1e-5,1e-2]$ and batch size in the range of $[32,256]$. The kernel size is 4 for all datasets. For CWGAN-GP, we used an Adam optimizer with a learning rate of 3e-4 and a batch size of 128. We train the model for 150 epochs. We use a gradient penalty coefficient ($\lambda_{GP}$) of 10 for CWGAN-GP. For Pix2Pix, we used an Adam optimizer with a learning rate of 2e-3 and a batch size of 32. We train the model for 150 epochs. The reconstruction loss coefficient ($\lambda_{L2}$) is 200 for P19 dataset and 100 for other datasets. We use a linear decay learning rate scheduler for the first 50 epochs. We use a gradient penalty coefficient ($\lambda_{GP}$) of 10 for CWGAN-GP.

\section{Metrics}
\label{app:metrics}

In this section, we describe the metrics used in the paper.

\subsection*{Precision, Recall, Density, Coverage}

Let X and Y be two sets of samples from the real and generated data distributions. \cite{kynkaanniemiImprovedPrecisionRecall2019} first proposed to construct the \textit{manifold} for  $P(X)$  and  $Q(Y)$  separately. This object is nearly identical to the probabilistic density function except that it does not sum to 1. Precision then measures the expected likelihood of fake samples against the real manifold and recall measures the expected likelihood of real samples against the fake manifold. Density improves upon the precision metric by ﬁxing the overestimation of the manifold around real outliers. Coverage improves upon the recall metric to better quantify this by building the nearest neighbor manifolds around the real samples, instead of the fake samples, as they have fewer outliers \cite{naeemReliableFidelityDiversity2020}.





\begin{align*}
    \text{Precision} & := \frac{1}{M} \sum_{j=1}^{M} \mathbb{I}_{y \in \text{manifold}(X_1, \ldots, X_N)},           \\
    \text{Recall}    & := \frac{1}{N} \sum_{i=1}^{N} \mathbb{I}_{x \in \text{manifold}(Y_1, \ldots, Y_M)},           \\
    \text{Density}   & := \frac{1}{kM} \sum_{j=1}^{M} \sum_{i=1}^{N} \mathbb{I}{y_j \in B(X_i, NND_k(X_i))},         \\
    \text{Coverage}  & := \frac{1}{N} \sum{i=1}^{N} \mathbb{I}_{\exists j \text{ s.t. } y_j \in B(X_i, NND_k(X_i))},
\end{align*}

where $N$ and $M$ are the number of real and generated samples.  $\mathbb{I}_{(\cdot)}$  is the indicator function, and manifolds are defined as:

\begin{equation*}
    \text{manifold}(X_1, \ldots, X_N) := \bigcup_{i=1}^{N} B(X_i, \text{NND}(X_i)),
\end{equation*}

where \( B(x, r) \) is the sphere in \( \mathbb{R}^D \) around \( x \) with radius \( r \). \( \text{NND}_k(X_i) \) denotes the distance from \( X_i \) to the \( k^{th} \) nearest neighbour among \( \{X_j\} \) excluding itself.


\subsection*{Nearest Neighbor Adversarial Accuracy (NNAA)}

This metric was proposed in \cite{yaleGenerationEvaluationPrivacy2020} to measure the degree to which a generative model overfits the real training data. The Adversarial Accuracy (AA) between two sets of samples X and Y is defined as:

\begin{equation*}
    AA(X,Y) = \frac{1}{2} \left( \frac{1}{N} \sum_{i=1}^{N} \mathbb{I}\{d_{XY}(i) > d_{XX}(i)\} + \frac{1}{N} \sum_{i=1}^{N} \mathbb{I}\{d_{YX}(i) > d_{YY}(i)\} \right),
\end{equation*}

where \( \mathbb{I}\{\cdot\} \) is the indicator function, and the distances are defined as:

\begin{align*}
    d_{XY}(i) = \min_{j} \lVert x_{X}^{(i)} - x_{Y}^{(j)} \rVert, \quad d_{YX}(i) = \min_{j} \lVert x_{Y}^{(i)} - x_{X}^{(j)} \rVert, \\
    \quad d_{XX}(i) = \min_{\substack{j, j \neq i}} \lVert x_{X}^{(i)} - x_{X}^{(j)} \rVert, \quad d_{YY}(i) = \min_{\substack{j, j \neq i}} \lVert x_{Y}^{(i)} - x_{Y}^{(j)} \rVert,
\end{align*}

In principle, we expect that the nearest neighbors of a sample in either set $X$ or $Y$ would be from the opposite set in 50 percent of cases. Thus, the ideal value of $AA(X,Y)$ is 0.5. However, if the model overfits the training data (i.e. the model memorizes the training data), the value of $AA(X,Y)$ will be lower than 0.5 because the nearest neighbors of a sample in $X$ or $Y$ would be from the opposite set in most cases. Thus, we can interpret $AA(X,Y)<0.5$ as a sign of overfitting or privacy issues. Considering three sets of samples $X_{train}$, $X_{test}$, and $X_{synthetic}$, we can define the the following metrics:

\begin{align*}
    AA_{train} = AA(X_{train}, X_{synthetic}), \\
    AA_{test} = AA(X_{test}, X_{synthetic}),   \\
    AA_{baseline} = AA(X_{train}, X_{test})    \\
    NNAA = AA_{\text{test}} - AA_{\text{train}},
\end{align*}

\section{Additional Results}
\label{app:results}

\subsection*{Statistical Similarity}

Tables \ref{tab:meanstd-p12}, \ref{tab:meanstd-p19} and \ref{tab:meanstd-mimic} show the statistical similarity for P12, P19, and MIMIC datasets, respectively. We report the mean and standard deviation of each variable for the train and test data as well as the generated data from TimEHR and RTSGAN. We use Cohen's d to quantify the effect size of the difference between two group means, indicating the magnitude of the observed difference in standard deviation units. We can see that TimEHR has a better mean and standard deviation for most of the variables especially for the variables with a high missing rate.It is notable that in MIMIC-III dataset, RTSGAN has failed to generate any data for three variables.

\begin{table}[htbp]
    \small
    \centering
    \caption{Statistical Similarity for P12 dataset. (*) and (**) denote medium and large effect size, respectively.}

\begin{tabular}{p{12.975em}p{5.605em}p{5.395em}ll}
\toprule
Variable (Missing Rate\%) & Train & Test  & \multicolumn{1}{p{6.13em}}{TimEHR} & \multicolumn{1}{p{6.13em}}{RTSGAN} \\
\midrule
HR (4.3) & 86.4 (17.3) & 86.3 (17.5) & 88.8 (12.4) & \multicolumn{1}{p{6.13em}}{89 (17.6)} \\
Urine (27.2) & 103 (99.9) & 108 (106) & 111 (80.3) & \multicolumn{1}{p{6.13em}}{116 (86.4)} \\
SysABP (42.6) & 120 (22.2) & 120 (22.2) & 118 (17.6) & \multicolumn{1}{p{6.13em}}{118 (18.5)} \\
DiasABP (42.8) & 59.4 (11.7) & 59.8 (11.9) & 60.9 (8.85) & \multicolumn{1}{p{6.13em}}{58.9 (9.39)} \\
MAP (43) & 79.7 (14.3) & 80.1 (14.5) & 79.6 (10.9) & \multicolumn{1}{p{6.13em}}{79 (11.2)} \\
ALP (98.3) & 108 (85.3) & 102 (70.5) & 97.4 (55.8) & \multicolumn{1}{p{6.13em}}{116 (74.4)} \\
Albumin (98.7) & 2.88 (0.647) & 2.91 (0.656) & 3 (0.574) & \multicolumn{1}{p{6.13em}}{3.31 (0.604)**} \\
TroponinT (98.9) & 0.814 (1.54) & 0.871 (1.58) & 0.787 (1.08) & \multicolumn{1}{p{6.13em}}{2.04 (2.25)**} \\
TroponinI (99.8) & 7.22 (9.04) & 7.54 (10.8) & 7.26 (6.52) & \multicolumn{1}{p{6.13em}}{2.16 (1.14)**} \\
Cholesterol (99.8) & 153 (40.9) & 154 (39.6) & 138 (37.4)* & 164 (28.2)* \\
\bottomrule
\end{tabular}%

    \label{tab:meanstd-p12}%

\end{table}%

\begin{table}[htbp]
    \small
    \centering
    \caption{Statistical Similarity for P19 dataset. (*) and (**) denote medium and large effect size, respectively.}

\begin{tabular}{p{12.975em}p{5.605em}p{5.395em}ll}
\toprule
Variable (Missing Rate\%) & Train & Test  & \multicolumn{1}{p{6.13em}}{TimEHR} & \multicolumn{1}{p{6.13em}}{RTSGAN} \\
\midrule
HR (2.3) & 84.2 (16.8) & 83.8 (16.6) & 81.4 (12.3) & \multicolumn{1}{p{6.13em}}{85 (15.8)} \\
MAP (5.4) & 81.9 (15.3) & 81.7 (15.1) & 80.5 (13.5) & \multicolumn{1}{p{6.13em}}{81.8 (14.2)} \\
O2Sat (6.1) & 97.3 (2.41) & 97.4 (2.39) & 97.1 (2.1) & \multicolumn{1}{p{6.13em}}{97.6 (1.93)} \\
SBP (7.4) & 123 (22.3) & 123 (22.3) & 123 (20.5) & \multicolumn{1}{p{6.13em}}{123 (19.9)} \\
Resp (8.7) & 18.5 (4.6) & 18.4 (4.57) & 18 (3.81) & \multicolumn{1}{p{6.13em}}{18.1 (3.92)} \\
AST (98.3) & 151 (342) & 150 (326) & 103 (238) & \multicolumn{1}{p{6.13em}}{144 (219)} \\
Bilirubin total (98.4) & 1.45 (1.85) & 1.4 (1.64) & 1.27 (1.33) & \multicolumn{1}{p{6.13em}}{1.51 (1.46)} \\
TroponinI (98.9) & 5.76 (13.1) & 5.73 (13.2) & 3.28 (6.59)* & \multicolumn{1}{p{6.13em}}{3.32 (5.14)*} \\
Fibrinogen (99.3) & 277 (131) & 275 (130) & 252 (94.3)* & \multicolumn{1}{p{6.13em}}{221 (93.2)*} \\
Bilirubin direct (99.8) & 1.3 (2.01) & 1.15 (1.94) & 1.01 (1.24) & 2.05 (2.21)* \\
\bottomrule
\end{tabular}%

    \label{tab:meanstd-p19}%

\end{table}%

\begin{table}[htbp]
    \small
    \centering
    \caption{Statistical Similarity for MIMIC-III dataset. (*) and (**) denote medium and large effect size, respectively. NaN indicates that no data has been generated for that variable.}

\begin{tabular}{p{12.975em}p{5.605em}p{5.395em}ll}
\toprule
Variable (Missing Rate\%) & Train & Test  & \multicolumn{1}{p{6.13em}}{TimEHR} & \multicolumn{1}{p{6.13em}}{RTSGAN} \\
\midrule
heartrate (3.2) & 85.2 (16.9) & 85.2 (16.9) & 85.9 (13.6) & \multicolumn{1}{p{6.13em}}{84.6 (16.3)} \\
resprate (5.3) & 19.3 (5.24) & 19.4 (5.27) & 19.5 (4.54) & \multicolumn{1}{p{6.13em}}{19.2 (4.68)} \\
meanbp (7) & 78.1 (14.2) & 78.2 (14.2) & 78.8 (11.9) & \multicolumn{1}{p{6.13em}}{78.2 (14.2)} \\
sysbp (7.1) & 120 (21) & 120 (20.9) & 122 (17.7) & \multicolumn{1}{p{6.13em}}{120 (19.4)} \\
diasbp (7.3) & 60 (13.3) & 60 (13.3) & 60.4 (11.6) & \multicolumn{1}{p{6.13em}}{60 (13)} \\
bg chloride (99.8) & 106 (5.94) & 107 (5.87) & 107 (5.92) & \multicolumn{1}{p{6.13em}}{104 (4.28)**} \\
bands (99.8) & 7.85 (7.94) & 7.87 (7.69) & 7.62 (6.18) & \multicolumn{1}{p{6.13em}}{9.53 (7.02)*} \\
bg methemoglobin (100) & 1.82 (3.51) & 3.2 (5.11)* & 7.49 (3.62)** & \multicolumn{1}{p{6.13em}}{NaN} \\
bg carboxyhemoglobin (100) & 1.23 (0.531) & 1.31 (0.73) & 1.5 (0.355)** & \multicolumn{1}{p{6.13em}}{NaN} \\
bg bicarbonate (100) & 23.4 (5.12) & 21.5 (4.46)* & 22.6 (4.99) & NaN \\
\bottomrule
\end{tabular}%

    \label{tab:meanstd-mimic}%

\end{table}%

\newpage



\subsection*{Image-based Visualization}

\begin{figure}[htbp]
    \centering
    \includegraphics[width=0.8\textwidth]{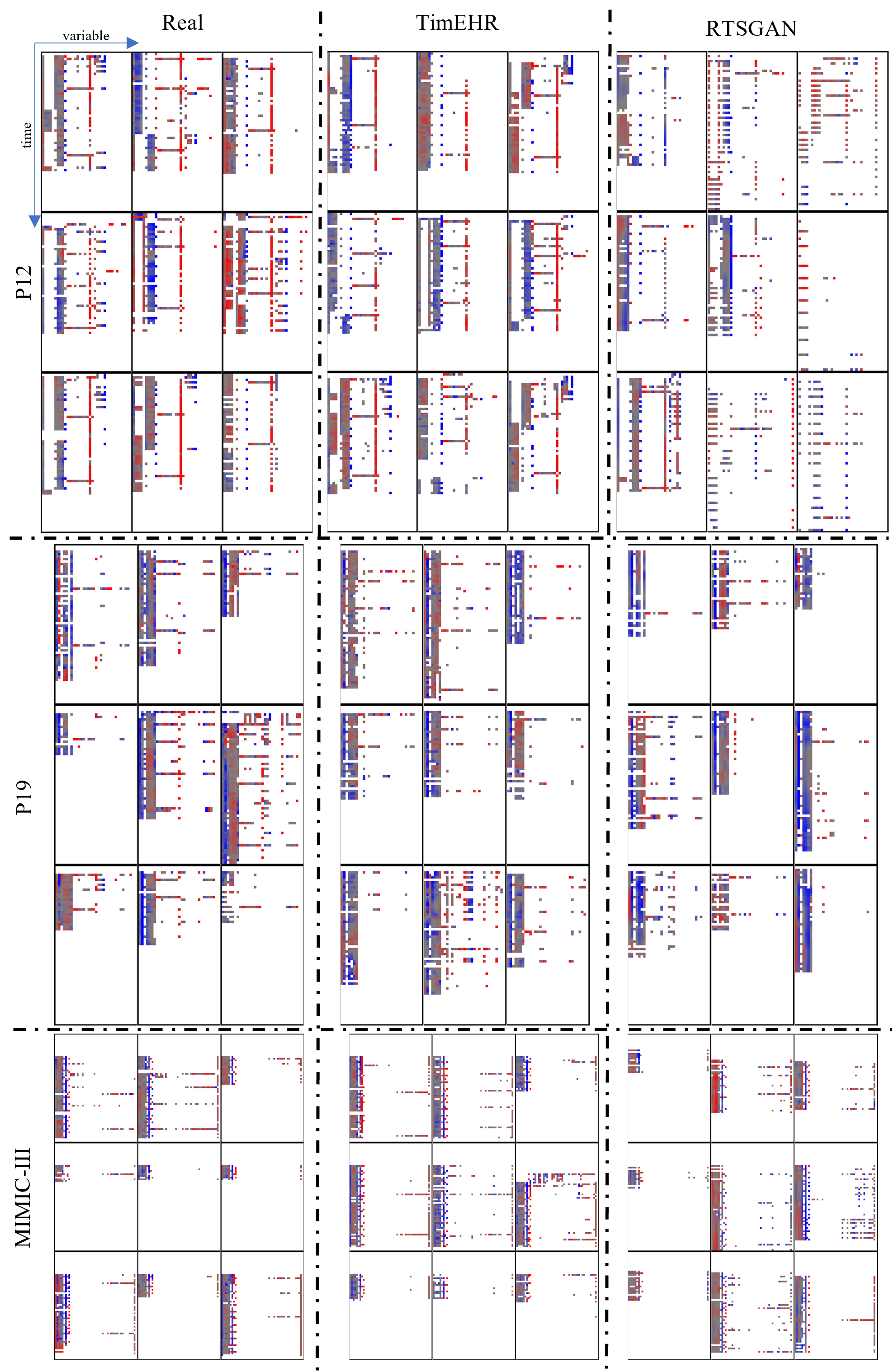}
    \caption{ Image-based visualization of real and generated time series for three EHR datasets.}
    \label{fig:vis}
\end{figure}

We visualize the image-based representation of a nine real and generated time series in \cref{fig:vis}. This representation has two advantages: (1) it can highlight the missingness pattern, and (2) it can show all variables in a single image. We observe that TimEHR samples are more realistic. In particular, the missingness pattern of RTSGAN samples are visually different from the real data.

\newpage

\subsection*{Correlation Matrix}

\begin{figure}[htbp]
    \centering
    \includegraphics[width=0.8\textwidth]{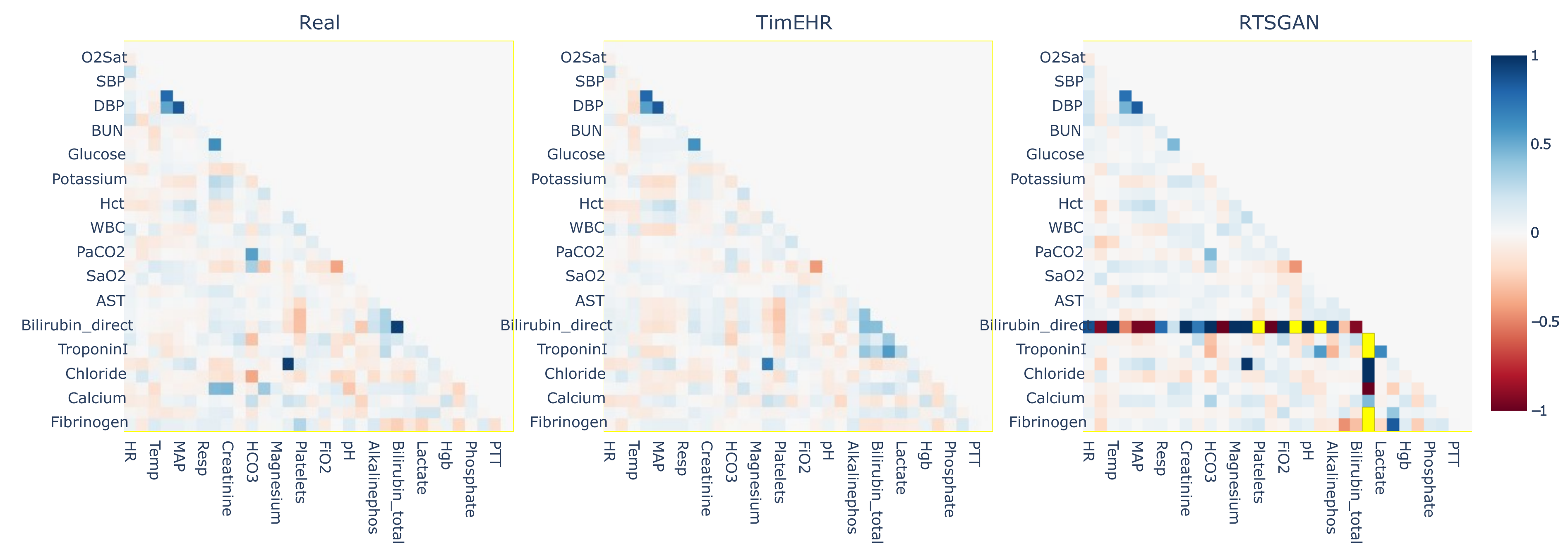}
    \caption{Correlation matrix for P19 dataset. The NaN values are shown in yellow.}
    \label{fig:corr-p19}
\end{figure}

\begin{figure}[htbp]
    \centering
    \includegraphics[width=0.8\textwidth]{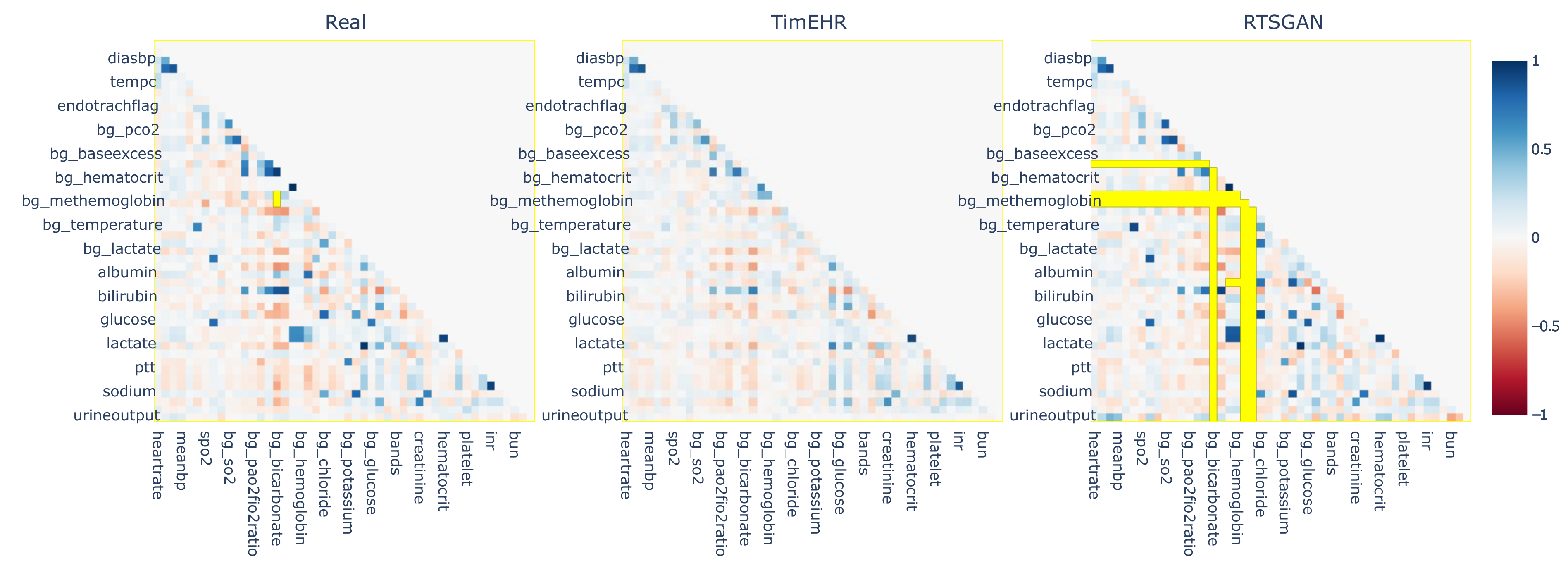}
    \caption{Correlation matrix for MIMIC-III dataset. The NaN values are shown in yellow.}
    \label{fig:corr-mimic}
\end{figure}

Figures \ref{fig:corr-p19}, and \ref{fig:corr-mimic} show the correlation matrix for  P19, and MIMIC-III datasets, respectively. We observe that TimEHR has a better correlation matrix compared to RTSGAN.

\subsection*{Ablation Study}
In this section, we report the absolute metric values from \cref{fig:ablation}. Tables \ref{tab:ablation-fidelity}, \ref{tab:ablation-utility}, and \ref{tab:ablation-privacy} show the fidelity, utility, and privacy metrics, respectively. We can see that \textit{TimEHR w/o CWGAN} has fidelity scores that are better than the baseline and privacy issues as well. The perfect utility score of \textit{TimEHR w/o CWGAN} is not desired as the model is memeorizing the training data.

\begin{table}[htbp]
    \small
    \centering
    \caption{Fidelity metrics in the ablation study.}

\begin{tabular}{ccccccc}
\toprule
Dataset & Model & precision ($\uparrow$) & recall ($\uparrow$) & density ($\uparrow$) & coverage ($\uparrow$) & TCD ($\downarrow$) \\
\midrule
\multirow{5}[2]{*}{P12} & TimEHR wo CWGAN & \textbf{1(0.0)} & \textbf{1.21(0.025)} & \textbf{1(0.0)} & \textbf{1(0.0)} & \textbf{0.004(0.001)} \\
      & TimEHR wo Pix2Pix & 0.392(0.049) & 0.309(0.074) & 0.431(0.053) & 0.212(0.037) & 0.072(0.006) \\
      & TimEHR wo L2 & 0.302(0.079) & 0.317(0.118) & 0.468(0.06) & 0.054(0.021) & 0.086(0.023) \\
      & TimEHR & 0.782(0.024) & 0.657(0.054) & 0.775(0.02) & 0.651(0.05) & 0.019(0.001) \\
      & BL    & \textit{0.967(0.006)} & \textit{0.94(0.051)} & \textit{0.855(0.015)} & \textit{0.854(0.006)} & \textit{0.01(0.001)} \\
\midrule
\multirow{5}[2]{*}{P19} & TimEHR wo CWGAN & \textbf{0.999(0.002)} & \textbf{1.43(0.044)} & \textbf{0.998(0.003)} & \textbf{0.984(0.03)} & \textbf{0.008(0.003)} \\
      & TimEHR wo Pix2Pix & 0.297(0.074) & 0.25(0.076) & 0.361(0.066) & 0.273(0.087) & 0.042(0.004) \\
      & TimEHR wo L2 & 0.159(0.048) & 0.206(0.066) & 0.416(0.055) & 0.047(0.014) & 0.091(0.051) \\
      & TimEHR & 0.77(0.051) & 0.814(0.104) & 0.791(0.026) & 0.657(0.03) & 0.021(0.002) \\
      & BL    & \textit{0.964(0.005)} & \textit{0.959(0.037)} & \textit{0.852(0.006)} & \textit{0.849(0.003)} & \textit{0.007(0.001)} \\
\midrule
\multirow{5}[2]{*}{MIMIC-III} & TimEHR wo CWGAN & \textbf{0.972(0.02)} & \textbf{1.19(0.136)} & \textbf{0.99(0.005)} & \textbf{0.947(0.017)} & \textbf{0.014(0.002)} \\
      & TimEHR wo Pix2Pix & 0.377(0.051) & 0.322(0.14) & 0.431(0.105) & 0.358(0.1) & 0.065(0.003) \\
      & TimEHR wo L2 & 0.171(0.035) & 0.24(0.085) & 0.369(0.086) & 0.036(0.011) & 0.124(0.022) \\
      & TimEHR & 0.798(0.055) & 0.759(0.111) & 0.762(0.027) & 0.622(0.039) & 0.035(0.001) \\
      & BL    & \textit{0.839(0.074)} & \textit{0.642(0.192)} & \textit{0.778(0.052)} & \textit{0.719(0.079)} & \textit{0.03(0.004)} \\
\bottomrule
\end{tabular}%

    \label{tab:ablation-fidelity}%

\end{table}%

\begin{table}[htbp]
    \small
    \centering
    \caption{Utility metrics in the ablation study.}

\begin{tabular}{cccc}
\toprule
Dataset & Model & AUROC & AUPRC \\
\midrule
\multirow{5}[2]{*}{P12} & TimEHR wo CWGAN & \textbf{0.826(0.012)} & \textbf{0.462(0.029)} \\
      & TimEHR wo Pix2Pix & 0.768(0.038) & 0.375(0.05) \\
      & TimEHR wo L2 & 0.654(0.043) & 0.265(0.036) \\
      & TimEHR & 0.785(0.015) & 0.424(0.035) \\
      & BL    & \textit{0.832(0.008)} & \textit{0.475(0.023)} \\
\midrule
\multirow{5}[2]{*}{P19} & TimEHR wo CWGAN & \textbf{0.888(0.01)} & \textbf{0.592(0.024)} \\
      & TimEHR wo Pix2Pix & 0.77(0.044) & 0.314(0.084) \\
      & TimEHR wo L2 & 0.602(0.048) & 0.113(0.02) \\
      & TimEHR & 0.853(0.014) & 0.452(0.017) \\
      & BL    & \textit{0.908(0.011)} & \textit{0.618(0.011)} \\
\midrule
\multirow{5}[2]{*}{MIMIC-III} & TimEHR wo CWGAN & \textbf{0.829(0.009)} & \textbf{0.553(0.025)} \\
      & TimEHR wo Pix2Pix & 0.8(0.013) & 0.493(0.031) \\
      & TimEHR wo L2 & 0.66(0.049) & 0.291(0.051) \\
      & TimEHR & 0.82(0.01) & 0.536(0.024) \\
      & BL    & \textit{0.863(0.012)} & \textit{0.621(0.028)} \\
\bottomrule
\end{tabular}%

    \label{tab:ablation-utility}%

\end{table}%

\begin{table}[htbp]
    \small
    \centering
    \caption{Privacy metrics in the ablation study. \textbf{Bold} values indicate privacy issues.}

\begin{tabular}{ccccc}
\toprule
Dataset & Model & $AUROC_{MIA}$ & $AA_{train}$ & $AA_{test}$ \\
\midrule
\multirow{5}[2]{*}{P12} & TimEHR wo CWGAN & \textbf{0.003(0.001)} & \textbf{0.001(0.001)} & 0.551(0.01) \\
      & TimEHR wo Pix2Pix & 0.497(0.009) & 0.935(0.009) & 0.937(0.013) \\
      & TimEHR wo L2 & 0.496(0.012) & 0.957(0.013) & 0.96(0.01) \\
      & TimEHR & 0.494(0.01) & 0.735(0.017) & 0.747(0.015) \\
      & BL    & \textit{0.5(0.0)} & \textit{0.512(0.01)} & \textit{0.512(0.01)} \\
\midrule
\multirow{5}[2]{*}{P19} & TimEHR wo CWGAN & \textbf{0.124(0.118)} & \textbf{0.067(0.102)} & 0.564(0.018) \\
      & TimEHR wo Pix2Pix & 0.5(0.006) & 0.936(0.022) & 0.941(0.018) \\
      & TimEHR wo L2 & 0.499(0.008) & 0.979(0.006) & 0.979(0.008) \\
      & TimEHR & 0.499(0.008) & 0.721(0.025) & 0.736(0.029) \\
      & BL    & \textit{0.5(0.0)} & \textit{0.518(0.012)} & \textit{0.518(0.012)} \\
\midrule
\multirow{5}[2]{*}{MIMIC-III} & TimEHR wo CWGAN & \textbf{0.223(0.043)} & \textbf{0.185(0.049)} & 0.781(0.076) \\
      & TimEHR wo Pix2Pix & 0.493(0.019) & 0.9(0.019) & 0.945(0.03) \\
      & TimEHR wo L2 & 0.495(0.027) & 0.977(0.006) & 0.988(0.006) \\
      & TimEHR & 0.488(0.024) & 0.733(0.041) & 0.837(0.061) \\
      & BL    & \textit{0.5(0.0)} & \textit{0.669(0.092)} & \textit{0.669(0.092)} \\
\bottomrule
\end{tabular}%

    \label{tab:ablation-privacy}%

\end{table}%

\end{document}